%% file: main.tex
\documentclass[]{fairmeta}

\input{math_commands.tex}
\usepackage{float}
\usepackage{graphicx}
\graphicspath{{figs/}}
\usepackage{multirow}
\usepackage[utf8]{inputenc} % allow utf-8 input
\usepackage[T1]{fontenc}    % use 8-bit T1 fonts
\usepackage{hyperref}       % hyperlinks
\usepackage{url}            % simple URL typesetting
\usepackage{booktabs}       % professional-quality tables
\usepackage{amsfonts}       % blackboard math symbols
\usepackage{nicefrac}       % compact symbols for 1/2, etc.
\usepackage{microtype}      % microtypography
\usepackage{xcolor}         % colors
\usepackage{tablefootnote}
\usepackage{xspace}
\usepackage{dsfont}

\usepackage{listings}
\lstdefinelanguage{yaml}{
  keywords={true,false,null,y,n},
  keywordstyle=\color{blue!70!black}\bfseries,
  sensitive=false,
  comment=[l]{\#},
  commentstyle=\color{gray}\itshape,
  stringstyle=\color{purple!70!black},
  morestring=[b]',
  morestring=[b]",
}
\lstdefinestyle{nbcode}{
  basicstyle=\scriptsize\ttfamily,
  breaklines=true,
  breakatwhitespace=true,
  columns=fullflexible,
  keepspaces=true,
  showstringspaces=false,
  xleftmargin=4pt,
  frame=none,
  upquote=true,
  literate={\$}{{\$}}1,
}
\newcommand{\ns}{NeuralSet\xspace}
\newcommand{\nt}{NeuralTrain\xspace}
\newcommand{\nf}{NeuralFetch\xspace}
\newcommand{\nb}{NeuralBench\xspace}
\newcommand{\nbe}{\texttt{NeuralBench-EEG v1.0}\xspace}
\newcommand{\nbecore}{\texttt{NeuralBench-EEG-Core v1.0}\xspace}
\newcommand{\nbefull}{\texttt{NeuralBench-EEG-Full v1.0}\xspace}
\newcommand{\nburl}{\url{https://github.com/facebookresearch/neuroai/tree/main/neuralbench-repo}\xspace}
\newcommand{\ntasks}{36\xspace}
\newcommand{\ndatasets}{94\xspace}
\newcommand{\nhours}{13,603\xspace}
\newcommand{\nsubjects}{9,478\xspace}
\newcommand{\nmodels}{14\xspace}
\title{\nb: A Unifying Framework to Benchmark NeuroAI Models}

\author[1]{Hubert Banville}
\author[1]{Stéphane d'Ascoli}
\author[1]{Simon Dahan}
\author[1]{Jérémy Rapin}
\author[1]{Marlène Careil}
\author[1]{Yohann Benchetrit}
\author[1,4]{Jarod Lévy}
\author[1]{Saarang Panchavati}
\author[1]{Antoine Ratouchniak}
\author[2,3]{Mingfang (Lucy) Zhang}
\author[1]{Elisa Cascardi}
\author[1]{Katelyn Begany}
\author[1]{Teon Brooks}
\author[1]{Jean-Rémi King}

\affiliation[1]{Brain \& AI team, Meta FAIR}
\affiliation[2]{École Normale Supérieure, Université PSL, CNRS}
\affiliation[3]{Hospital Foundation Adolphe de
Rothschild}
\affiliation[4]{MIND, Inria}

\correspondence{\email{hubertjb@meta.com}, \email{jeanremi@meta.com}}

\metadata[Code]{\nburl}

\abstract{
Deep learning and large public datasets have recently catalyzed the proliferation of AI models for processing brain recordings.
However, systematically evaluating these models remains a challenge: not only do the preprocessing pipelines, training and finetuning approaches largely vary across studies, but their downstream evaluation is often limited to small sets of tasks and/or datasets.
Here, we present \nb: a unified framework for benchmarking AI models of brain activity. We accompany this framework with \nbe ~-- a large EEG benchmark that includes \ntasks electroencephalography (EEG) tasks and \nmodels deep learning architectures, and is evaluated on \ndatasets datasets accessed through a standardized interface.
This first EEG-focused release already highlights two main findings.
First, current foundation models only marginally outperform task-specific models.
Second, a large set of tasks (\eg cognitive decoding, clinical predictions) remain highly challenging, even for the best models.
Critically, \nb is designed for the integration of new tasks, datasets, models, and neuroimaging modalities, as illustrated by preliminary extensions to MEG and fMRI datasets and models.
Through this white paper, we invite the community to expand this open-source framework and work together toward a unified benchmarking standard for neuroimaging models.
}

\begin{document}

\maketitle

\begin{figure}[h!]
    \centering
    \includegraphics[width=\linewidth]{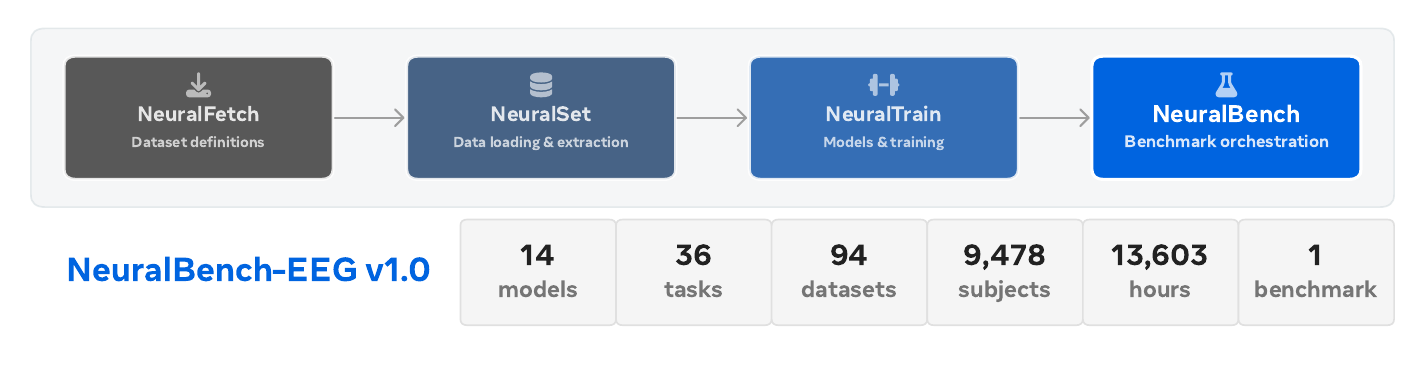}
    \caption{
    \nb is a benchmarking framework for the evaluation of brain models, based on the \nf, \ns and \nt libraries, and providing a fully configurable and flexible interface.
    \textbf{Second row.} Contents of the first release, \nbe.
    }
    \label{fig:device_overview}
\end{figure}

\begin{figure}
    \centering
    \includegraphics[width=1.0\linewidth]{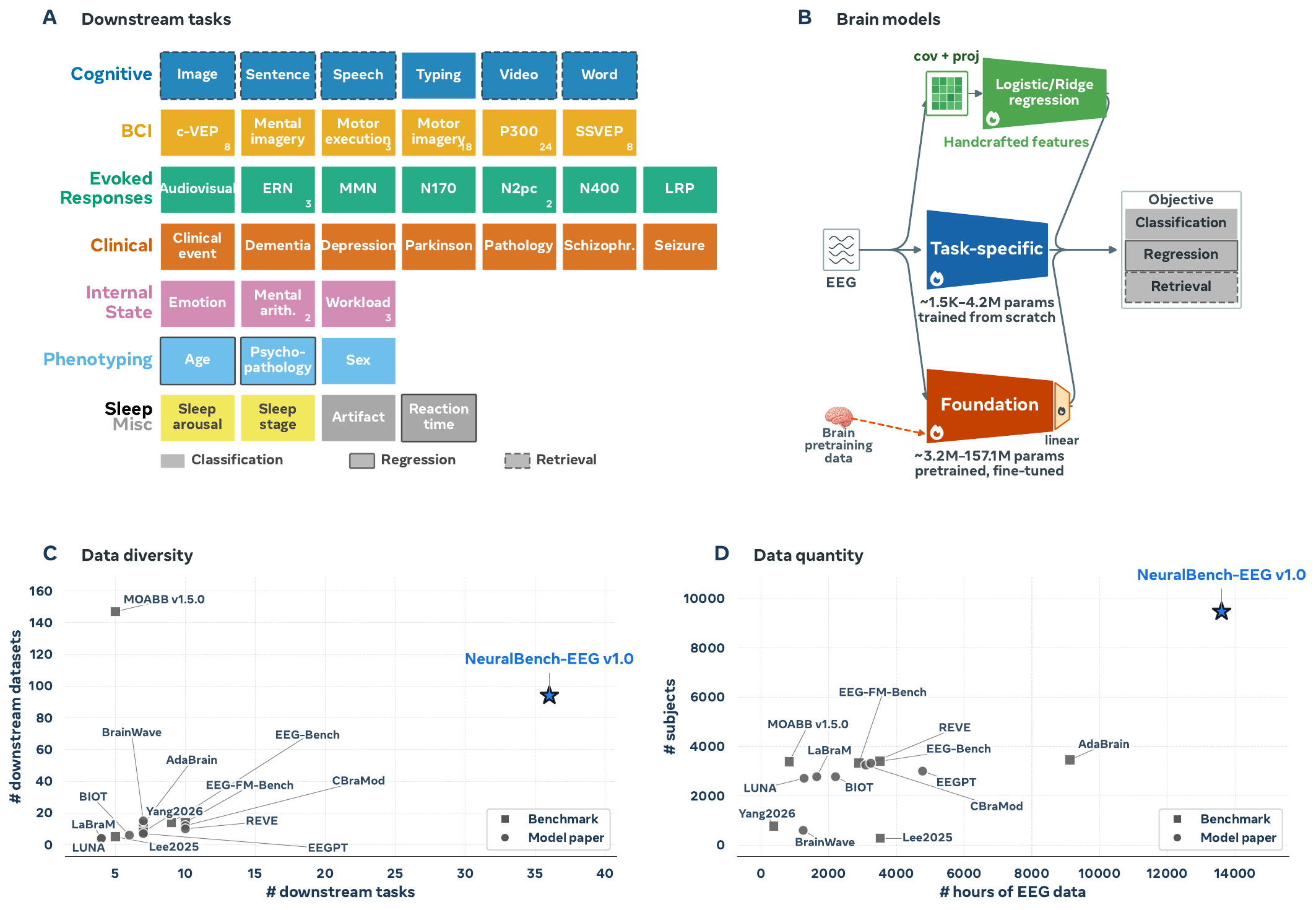}
    \caption{Detailed overview.
    \textbf{A}. \nbe currently includes \ntasks decoding downstream tasks on \ndatasets different datasets, covering 8 different categories such as cognitive decoding, brain-computer interfacing (BCI), evoked responses and clinical tasks.
    \textbf{B}. Three types of models are compared: task-specific from-scratch deep learning architectures, pretrained foundation models and handcrafted features-based baselines.
    \textbf{C-D}. \nbe currently contains more than three times the number of tasks of previous EEG downstream task benchmarks, and more than twice as many individual subjects, making it the largest such open-source benchmark yet.
    }
    \label{fig:overview}
\end{figure}

\section{Introduction}

\paragraph{The rise of AI models in neuroimaging.}
Historically focused on targeted, small-scale, and private studies \citep{poldrack2011inferring,michel2012towards,baillet2017magnetoencephalography},
the analysis of brain activity is now rapidly shifting toward large-scale public datasets~\citep{van2013wu,harati2014tuh,taylor2017cambridge,pinho2018individual,sun2025harvard,dascoli2026foundation}.
On top of boosting statistical power and stimulating reproducible research \citep{button2013power,poldrack2017scanning,dascoli2026foundation}, the proliferation of open datasets has catalyzed the use of AI to process brain recordings.
Specifically, the self-supervised learning methods developed for language, speech and images~\citep{devlin2019bert,baevski2020wav2vec,oquab2024dinov2} are now adapted for the development of 
\textit{brain foundation models}~\citep{yang2025foundation,zhou2025brain}.
These models start to transfer to downstream tasks spanning electroencephalography (EEG)~\citep{jiang2024labram,kostas2021bendr,yang2023biot,wang2024eegpt,doner2025luna,elouahidi2025reve}, functional magnetic resonance imaging (fMRI) ~\citep{kim2023swift,dong2024brainjepa,caro2023brainlm},
intracranial EEG~\citep{wang2023brainbert,yuan2024brainwave}, structural MRI~\citep{tak2026generalizable},
electrophysiology~\citep{azabou2023unified}, and calcium imaging~\citep{wang2025foundation}.

\paragraph{Current benchmarking challenges.}
Despite these advances, the evaluation landscape is fragmented.
For EEG, the Mother of All BCI Benchmarks (MOABB)\footnote{\url{https://moabb.neurotechx.com/}}~\citep{jayaram2018moabb,moabb2025} provides an extensive benchmark for brain-computer interfacing (148 datasets). However, it focuses on a narrow set of 5 downstream tasks (e.g., left versus right motor imagery) and primarily targets handcrafted features-based approaches, such as Riemannian geometry pipelines for trial classification~\citep{barachant2011multiclass,yger2016riemannian}.
Building on this, \citet{borra2024speechbrain} adapted the benchmark for deep learning, focusing specifically on a subset of nine BCI dataset from the MOABB and three well-established architectures.
In parallel, benchmarks tailored to EEG \emph{foundation} models have emerged, \eg, EEG-Bench~\citep{kastrati2025eeg}, EEG-FM-Bench~\citep{xiong2025eeg}, and AdaBrain-Bench~\citep{wu2025adabrain}. However, these efforts too are limited to a restricted set of downstream evaluations (\Cref{fig:overview}).
This fragmentation is even more pronounced in other brain recording modalities. For intracranial EEG (iEEG), \citet{karpowicz2024few} introduced FALCON, a few-shot evaluation benchmark for invasive BCI tasks, but is limited to 5 evaluation datasets.
Elsewhere, foundation models for fMRI, MEG, and other modalities are still evaluated on handpicked datasets without a unified, systematic benchmark.
Overall, there is currently no comprehensive framework for standardizing and efficiently evaluating brain foundation models across neuroimaging modalities.

\paragraph{Contribution.}
Here, we introduce \nb, a community-driven benchmarking framework for evaluating foundation models on brain activity recordings. \nb is designed to be:
\begin{enumerate}
    \item \textbf{Comprehensive}. As many relevant downstream tasks as possible should be included, for as many relevant brain imaging devices as possible. This helps properly support claims of models being truly foundational and provides a more nuanced assessment of the strengths and weaknesses of different models.
    \item \textbf{Rigorously validated}. Publicly available datasets often require careful curation and tailored onboarding efforts so their contents can be used properly. Similarly, the framing of learning tasks requires special care to avoid common pitfalls (\eg leakage, overfitting, etc.) and to be maximally informative. This also applies to foundation models -- each one is trained using a specific preprocessing pipeline that needs to be replicated to allow fair evaluation.
    \item \textbf{Standardized}. By providing a unified point-of-entry for diverse brain imaging devices (EEG, iEEG, MEG, fMRI, fNIRS, etc.), \nb makes it easy to build and evaluate multimodal pipelines, bringing us closer to integrated brain modeling.
    \item \textbf{Flexible}. \nb is configurable at all levels of the neuroimaging modeling pipeline (data source, preprocessing, architecture, optimization, metrics, etc.), offering a compact and flexible interface for running and evaluating brain models.
\end{enumerate}

The first release of \nb, \nbe, supports \ntasks tasks evaluated on \ndatasets EEG datasets (with some tasks spanning multiple datasets), accessed through a standardized interface.
This effort, which capitalizes on the careful curation of public datasets, provides unique support for cognitive decoding tasks (\ie image, sentence, speech, typing, video and word decoding), which, apart from image decoding, have been omitted in previous benchmarks.
In addition, we demonstrate the versatility of this framework and seed its expansion towards other brain imaging modalities by including a couple of illustrative MEG and fMRI tasks and models.

\paragraph{Goal and overview.}
In the following sections, we present a technical overview of \nb, followed by initial results on \nbe which highlight two main findings: current EEG foundation models only marginally outperform task-specific baselines, and many complex tasks (\eg, cognitive decoding, clinical prediction) remain challenging even for foundation models.
We then show how \nb seamlessly extends to MEG and fMRI modalities.

\paragraph{Toward a Community-Driven Standard.}
Recognizing that a unified standard is only as strong as the community behind it, we provide the \nb codebase\footnote{\nburl} as a starting point for global collaboration. We invite researchers across tasks and modalities to co-author its evolution and join us in defining the future of brain model evaluation.

\section{Methods}

In this section, we give a technical overview of \nb, including its general architecture, a description of supported downstream tasks, model architectures, adaptation strategies, and evaluation protocols.
We provide a high-level summary here and refer the reader to the package documentation for additional technical details.

\subsection{Benchmark architecture}

\nb is built on top of three main Python packages, which provide an interface with the data, processing and modeling tools that already exist in the neuroscience software ecosystem.
Specifically, public neuroimaging datasets are accessed through \nf, which fetches curated datasets from public repositories like OpenNeuro\footnote{\url{https://openneuro.org/}}, 
DANDI\footnote{\url{https://dandiarchive.org/}}, 
and NEMAR\footnote{\url{https://nemar.org/}}.
Then, neural data is prepared and made available as PyTorch-ready dataloaders with \ns -- a package designed to efficiently leverage existing packages like MNE-Python~\citep{Gramfort:2013}\footnote{\url{https://mne.tools/}} and nilearn~\citep{abraham2014machine}\footnote{\url{https://nilearn.github.io/}} for applying classic neuroimaging preprocessing pipelines, as well as HuggingFace\footnote{\url{https://huggingface.co/}} for extracting AI model embeddings (\eg on word, speech and image modalities).
Finally, \nt provides modular training code with PyTorch-Lightning\footnote{\url{https://lightning.ai}}, Pydantic\footnote{\url{https://pydantic.dev/}} and exca~\citep{exca}\footnote{\url{https://github.com/facebookresearch/exca}}(see~\Cref{fig:overview}D).

Once installed, \nb is run with a command-line interface (CLI). For example, the following can be used to run the \texttt{audiovisual stimulus} classification task on the MNE-Python sample dataset~\citep{Gramfort:2013}:

\begin{tcbraster}[raster columns=2, raster equal height, raster column skip=2mm, raster force size=false]
\begin{tcolorbox}[title={1. Install \nb}, width=0.38\linewidth]
% \begin{minted}[autogobble]{console}
\begin{lstlisting}[language=bash]
$ pip install neuralbench
$ neuralbench --help  # Trigger interactive path configuration
\end{lstlisting}
% \end{minted}
\end{tcolorbox}
\begin{tcolorbox}[title={2. Run the \texttt{audiovisual stimulus} task on EEG}, width=0.62\linewidth - 2mm]
% \begin{minted}[autogobble]{console}
\begin{lstlisting}[language=bash]
$ neuralbench eeg audiovisual_stimulus --download  # Download data
$ neuralbench eeg audiovisual_stimulus --prepare    # Prepare cache
$ neuralbench eeg audiovisual_stimulus                 # Run the task
\end{lstlisting}
% \end{minted}
\end{tcolorbox}
\end{tcbraster}

On top of allowing quick sanity-checking of the codebase, this small dataset ($<$300 examples) is an interesting case study for very-low-data regime downstream tasks.

\paragraph{Task configuration} Tasks are configured through lightweight YAML files, which describe  data sources, train/validation/test splits, neural data preprocessing (filtering, scaling, windowing, etc.), target processing if any (\eg extracting dense representations of rich stimuli), trainer configuration (number of epochs, learning rate, etc.), and metrics.

\begin{tcolorbox}[title={Example YAML configuration for \texttt{audiovisual stimulus} classification on EEG.
}]
% \begin{minted}{yaml}
\begin{lstlisting}[language=yaml]
data:
  study:
    source.name: Mne2013SampleEeg
    split:
      name: SklearnSplit
      valid_split_ratio: 0.2
      test_split_ratio: 0.2
      stratify_by: description
  neuro.baseline: [0.0, 0.2]
  target:
    name: LabelEncoder
    event_types: Stimulus
    event_field: description
    return_one_hot: true
  trigger_event_type: Stimulus
  start: -0.2
  duration: 1.0
loss.name: CrossEntropyLoss
metrics: BalancedAcc
\end{lstlisting}
\label{listing:task_config_yaml}
\end{tcolorbox}

\paragraph{Model configuration} Similarly, deep learning architectures are defined through YAML files that configure \nt models.
Thanks to its configurable pydantic layer, \nt can transparently access models from other libraries such as BrainDecode\footnote{\url{https://braindecode.org/}}~\citep{schirrmeister2017deep,braindecode2026}.

\subsection{Task and dataset selection}
A primary objective of \nb is to provide a wide coverage of downstream tasks.
Focusing on EEG for this first release, we selected a set of \ntasks representative downstream tasks, covering 8 categories: cognitive decoding, brain-computer interfacing (BCI), evoked responses, clinical, internal state, sleep, phenotyping and miscellaneous (see~\Cref{fig:overview}A). 
Note that each task does not necessarily fall neatly into one of these categories, \eg the \texttt{P300} classification task could also fall under the \textit{Evoked responses} category, however since it is usually collected as part of a P300 speller protocol we file it under \textit{BCI}. Similarly, the \texttt{psychopathology} classification task could go under the \textit{Clinical} category, but we opt to keep it under the same \textit{Phenotyping} category as other tasks based on \citet{shirazi2024hbn}. Nevertheless, we keep this operational organization to facilitate interpretation.

Several inclusion criteria were considered. First, whenever possible, openly accessible datasets, rather than datasets requiring additional manual steps, were prioritized.
Second, as several EEG datasets are approaching saturation (\eg binary \texttt{pathology detection} on the TUAB dataset~\citep{lopez2015automated}), we devoted special efforts to include challenging tasks on which performance is likely to be significantly improvable.
To this end, we include multiple \textit{cognitive decoding} tasks, where the goal is to recover a dense representation of the stimulus presented to participants, \eg speech, images, words and videos~\citep{defossez2022decoding,benchetrit2024brain,dascoli2025towards}.
Finally, we re-framed common EEG classification tasks, \ie \texttt{clinical event} classification on the TUEV dataset~\citep{Harati2015} and \texttt{artifact} classification on the TUAR dataset~\citep{hamid2020temple}, as \textit{multilabel} classification tasks to take into account that that some events can overlap.

\textit{Note on pretraining data overlap.} Existing EEG foundation models are trained on custom corpora curated by the model authors. As there is a limited set of publicly available EEG datasets, some of these models saw (part of) the downstream datasets during pretraining. We flag these instances (\eg with hashed bars in \Cref{fig:results_bar_plot}), rather than discard them, as we did not notice a clear trend suggesting pretraining ``leakage'' improves downstream performance on the same data. 

Finally, while we prioritized breadth over depth in our selection of datasets (\ie most tasks rely on a single core dataset), we also leveraged readily available datasets through MOABB\footnote{\url{https://moabb.neurotechx.com/}}~\citep{moabb2025} to offer multiple comparison points on a few tasks, such as \texttt{motor imagery} and \texttt{P300} decoding.
This gives rise to two benchmark variants: \nbecore, which focuses on a single dataset per task, and \nbefull, which has up to 24 datasets per task and therefore enables the study of within-task variability.

In total, \nbe covers \ntasks tasks, \ndatasets datasets, \nsubjects subjects and \nhours hours (see~\Cref{tab:datasets}).
Of note, no data is contained in the released package; datasets are accessed through the standardized interface provided by \ns and \nf~\citep{king2026neuralset}.
Also, all datasets sources already provide de-identified data; as a result, the benchmark only processes de-identified subject IDs.
More details on datasets and downstream tasks can be found in Appendix~\ref{app:dataset_details} and in the benchmark documentation\footnote{\nburl}.

\begin{table}
\caption{Tasks and datasets in \nbe. For each task, we indicate the \textit{Core} dataset name, the machine learning objective (\textit{reg}: regression, \textit{clf}: classification, \textit{retr}: retrieval), the number of unique EEG channels, the number of unique subjects, the dimensionality of the target (\ie the number of outputs), and the number of additional datasets in the \textit{Full} variant, if any.}
\label{tab:datasets}
{\small
\begin{tabular}{p{4cm} l c r r r r r}
\toprule
Task & Core dataset & Obj. & \#Ch. & \#Subj. & \#Hours & \#Out. & +Datasets \\
\midrule
Age & Shirazi2024 & reg & 129 & 2,858 & 95.3 & 1 &  \\
Artifact & Hamid2020 & clf & 26 & 213 & 300.2 & 5 &  \\
Audiovisual stimulus & MneSample2013 & clf & 60 & 1 & 0.1 & 4 &  \\
Clinical event & Harati2015 & clf & 21 & 370 & 20.1 & 6 &  \\
C-VEP & Thielen2021 & clf & 8 & 30 & 50.0 & 20 & 7 \\
Dementia diagnosis & Miltiadous2023 & clf & 19 & 88 & 19.5 & 3 &  \\
Depression diagnosis & Mumtaz2018 & clf & 19 & 64 & 20.4 & 2 &  \\
Emotion & Chen2023 & clf & 32 & 123 & 28.7 & 9 &  \\
Error-related negativity & Kappenman2021Ern & clf & 30 & 40 & 4.4 & 2 & 2 \\
Image & Gifford2022 & retr & 63 & 10 & 230.7 & 1,536 &  \\
Mental arithmetic & Zyma2019 & clf & 20 & 35 & 2.3 & 2 & 1 \\
Mental imagery & Scherer2015 & clf & 30 & 9 & 3.9 & 5 &  \\
Mental workload & Hinss2023 & clf & 63 & 29 & 21.4 & 3 & 2 \\
Mismatch negativity & Kappenman2021Mmn & clf & 30 & 40 & 11.1 & 2 &  \\
Motor execution & Srisrisawang2024 & clf & 60 & 20 & 16.0 & 16 & 2 \\
Motor imagery & Stieger2021 & clf & 60 & 62 & 141.5 & 4 & 17 \\
N170 & Kappenman2021N170 & clf & 30 & 40 & 3.6 & 2 &  \\
N2pc & Kappenman2021N2pc & clf & 30 & 40 & 3.6 & 2 & 1 \\
N400 & Kappenman2021N400 & clf & 30 & 40 & 1.3 & 2 &  \\
P300 & Schreuder2010 & clf & 60 & 21 & 70.5 & 2 & 23 \\
Parkinsons diagnosis & Singh2021 & clf & 60 & 129 & 54.9 & 2 &  \\
Pathology & Lopez2017 & clf & 21 & 2,329 & 971.7 & 2 &  \\
Psychopathology & Shirazi2024 & reg & 129 & 2,820 & 94.0 & 1 &  \\
Reaction time & Shirazi2024 & reg & 129 & 1,945 & 70.8 & 1 &  \\
Lateralized readiness potential & Kappenman2021Lrp & clf & 30 & 40 & 4.4 & 2 &  \\
Schizophrenia diagnosis & Albrecht2019 & clf & 62 & 77 & 34.6 & 2 &  \\
Seizure & Dan2023 & clf & 24 & 23 & 982.9 & 2 &  \\
Sentence & Hollenstein2018 & retr & 128 & 12 & 21.5 & 768 &  \\
Sex & Shirazi2024 & clf & 129 & 2,858 & 95.3 & 2 &  \\
Sleep arousal & Ghassemi2018 & clf & 6 & 994 & 7,662.3 & 2 &  \\
Sleep stage & Kemp2000 & clf & 2 & 78 & 1,635.2 & 5 &  \\
Speech & Brennan2019 & retr & 60 & 33 & 58.5 & 4,096 &  \\
Typing & Levy2025 & clf & 61 & 20 & 40.7 & 29 &  \\
Video & Liu2024 & retr & 62 & 20 & 16.3 & 1,408 &  \\
Steady-state visually evoked potential & Wang2017 & clf & 64 & 34 & 9.1 & 40 & 6 \\
Word & Nieuwland2018 & retr & 74 & 222 & 441.4 & 1,024 &  \\
\bottomrule
\end{tabular}
}
\end{table}

\subsection{Model architecture selection}
We selected commonly used EEG architectures and foundation models.
First, we included 8 task-specific neural network architectures (\Cref{tab:classic_models}), \ie models that are trained from scratch on downstream tasks, typically with lower parameter counts.
These include
ShallowFBCSPNet and Deep4Net~\citep{schirrmeister2017deep},
EEGNet~\citep{lawhern2018eegnet}, BDTCN~\citep{gemein2020bdtcn},
ATCNet~\citep{altaheri2022atcnet}, EEGConformer~\citep{song2022eegconformer}, SimpleConvTimeAgg~\citep{elouahidi2023eegsimpleconv} and
CTNet~\citep{zhao2024ctnet}.

Second, we selected 6 EEG foundation models (\Cref{tab:foundation_models}), \ie larger models that have been pretrained on unlabelled data, and that are typically finetuned on downstream tasks.
These are BENDR~\citep{kostas2021bendr}, LaBraM~\citep{jiang2024labram}, BIOT~\citep{yang2023biot}, CBraMod~\citep{wang2025cbramod},  LUNA~\citep{doner2025luna} and REVE~\citep{elouahidi2025reve}.

Third, we included representative handcrafted features-based models using sklearn-like pipelines to provide a non-deep learning comparison point (see \Cref{tab:feature_based_baselines}).
These pipelines all rely on symmetric positive definite (SPD) matrix representations of the input EEG which are fed to `shallow` predictors (logistic or Ridge regression).

Finally, we include chance-level and ``dummy'' baselines to offer a proper performance floor to which models can be compared.
Chance-level performance corresponds to the performance obtained by an untrained, randomly initialized model.
``Dummy'' performance corresponds to the performance obtained by a model that predicts either the average target for regression and retrieval tasks, the most frequent class for binary and multiclass classification tasks, or samples class-wise labels based on training set statistics for multilabel classification tasks.
The dummy models are always fitted on the same training and validation sets as the other models.

\subsection{Downstream strategy protocol}

The original publications presenting the EEG foundation models listed above use different adaptation recipes, \ie for optimization (batch size, optimizer, learning rate, scheduler, etc.), aggregation (\texttt{[CLS]} token, average patch token, etc.) and projection head design (linear layer, MLP, etc.).
Here, to focus the comparison on model architecture and pretraining methodology, we adopt a shared training recipe.

Each foundation model is finetuned end-to-end, using a linear projection head which receives the average-pooled output tokens and outputs a single vector matching the target dimension.
Training uses AdamW~\citep{loshchilov2017decoupled} with a learning rate of $10^{-4}$, weight decay of 0.05, and cosine-annealing (10\% warmup), for up to 50 epochs with early stopping on the corresponding validation metric (patience=10).
The sole exception to this unified recipe is BENDR~\citep{kostas2021bendr}, for which we lower the learning rate to $10^{-5}$ and apply gradient clipping at 0.5 to obtain stable learning curves.
Therefore, model-specific techniques from the original papers — such as layer-wise learning rate decay, two-stage probing, or LoRA~\citep{hu2022lora} — are intentionally omitted and the comparison of downstream adaptation strategies is kept for future iterations of \nb.
Finally, task-specific architectures are trained \textit{from scratch} using randomly initialized weights, using the same optimization recipe as the foundation models.

\subsection{Preprocessing}

We take particular care to follow the preprocessing pipelines described in the original EEG foundation model papers, as pretrained checkpoints expect specific input statistics which require identical preprocessing.
For task-specific architectures, we use a single preprocessing configuration inspired by previous EEG decoding work~\citep{defossez2022decoding,benchetrit2024brain}: resampling to 120~Hz, bandpass filtering between 0.1 and 75~Hz, notch filters at 50 and 60~Hz as well as their harmonics, channel-wise robust scaling at the recording level, followed by clamping at 20. Nevertheless, it will be important, in future iterations, to systematically evaluate the impact of preprocessing strategies on downstream evaluation. 

\subsection{Data Splitting}
The generalization capacity of a model can be highly dependent on the task. Consequently, to split datasets into training, validation and testing sets, we use different strategies for each task: (1) predefined splits when provided by the authors, (2) ``leave-concept-out'' for cognitive decoding tasks, (3) cross-subject splits, with optional stratification for clinical tasks, or (4) random splits when very few examples are available.
We train (or finetune) each model on each task three times, using a single train/valid/test partition but three different random seeds for the initialization of the (non-pretrained) model weights.
See Appendix~\ref{sec:splitting} for more details.

\subsection{Training and evaluation metrics}
The training, finetuning and evaluation of the models must vary depending on the task objective.
In binary and \textit{multiclass} classification tasks, we use cross-entropy with optional weighing to mitigate class imbalance.
In \textit{multilabel} classification tasks, we instead use binary cross-entropy per target.
Regression tasks rely on the mean squared error (MSE) loss.
Finally, we also implemented retrieval tasks, where a model must recover which candidate target out of a given retrieval set actually corresponds to the input (\eg which image out of a set of multiple images seen by a participant, based on an input EEG window). Retrieval is a particularly well-posed problem for cognitive tasks, as the content of an image, text or video is neither categorical, nor necessarily perfectly specified in a pretrained embedding. Retrieval ultimately allows future models and approaches to be compared with one another, with little assumption on how to represent the target.  
For retrieval, we here rely on the CLIP loss~\citep{radford2021learning}, using the configuration of \citet{defossez2022decoding}, \ie with the brain-to-target term only and a fixed temperature parameter $\tau=1$:

\begin{equation}
\label{eq:clip_loss}
    \mathcal{L}_{CLIP}(\theta) = -\frac{1}{B} \sum_{i=1}^{B} \log \frac{\exp(s(\hat{\bm{z}}_i, \bm{z}_i)/\tau)}{\sum_{j=1}^{B} \exp(s(\hat{\bm{z}}_i, \bm{z}_j)/\tau)}
\end{equation}

where $B$ is the batch size, $s$ is the cosine similarity, and $\bm{z}_i$ and $\hat{\bm{z}}_i$ are the target dense embedding (\eg DINOv2~\citep{oquab2024dinov2} for \texttt{image} retrieval) and corresponding prediction for batch element $i$.

\subsection{Model comparison}
To help summarize the comparison across models, we opted to report a single representative metric per task type: \ie
balanced accuracy for binary and multiclass classification, F1-score for multilabel classification, Pearson correlation for regression, and top-5 accuracy for retrieval\footnote{We evaluate top-5 accuracy on the entire test set, using within-subject aggregation of repeated predictions, if available.}.
See Appendix~\ref{sec:metrics} for more details.
We further report normalized scores $\tilde{s}$ to facilitate cross-task comparisons and emphasize task difficulty:

\begin{equation}
\label{eq:normalized_score}
    \tilde{s} = \frac{s - s_{\text{dummy}}}{s_{\text{perfect}} - s_{\text{dummy}}}
\end{equation}

where $s_{\text{dummy}}$ is the performance obtained by the dummy model on a specific dataset, and $s_{\text{perfect}}$ is the metric-specific theoretical highest performance obtainable (\eg 100\% balanced accuracy for classification tasks). 
Of note, due to the inherent noise in brain activity recordings, this perfect score cannot be attained in practice.
With this formulation, a value of 0 indicates dummy-level performance and a value of 1 indicates perfect performance.

Finally, global model rankings consists of the per-task rank of each model, averaged over all tasks.
On \nbefull, we additionally use all available datasets to compute this ranking, \ie we average ranks across datasets of the same task first.

\subsection{Computational considerations}

\nbefull requires running 4,947 experiments. 
(\nmodels models + chance, dummy \& handcrafted features baselines, 97 tasks-datasets, 3 seeds each).
By default, each job uses a single GPU with at least 32 GB VRAM and 64 GB of CPU RAM, though measured peak GPU usage averages only $\sim$1.3 GB (max $\sim$30.3 GB).
The full EEG benchmark requires $\sim$11 TB of disk space ($\sim$3.2 TB for the raw data, 7.8 TB of preprocessed cache, and 333 GB for logged results).
Training times vary widely, with a median of 2.7 minutes but mean of 21.7 minutes per experiment, with the longest run taking 18.2 hours.
This yields an estimated total serial runtime of $\sim$73 days (1,751 GPU-hours).
The benchmark runs on Python >= 3.12 with PyTorch 2.6~\citep{paszke2019pytorch} and Pytorch-Lightning~\citep{falcon2019pytorchligthning}.

\section{Results}

\subsection{NeuralBench-EEG v1.0: Evaluating EEG foundation models on \ntasks diverse downstream tasks}
\begin{figure}
    \centering
    \includegraphics[width=\linewidth]{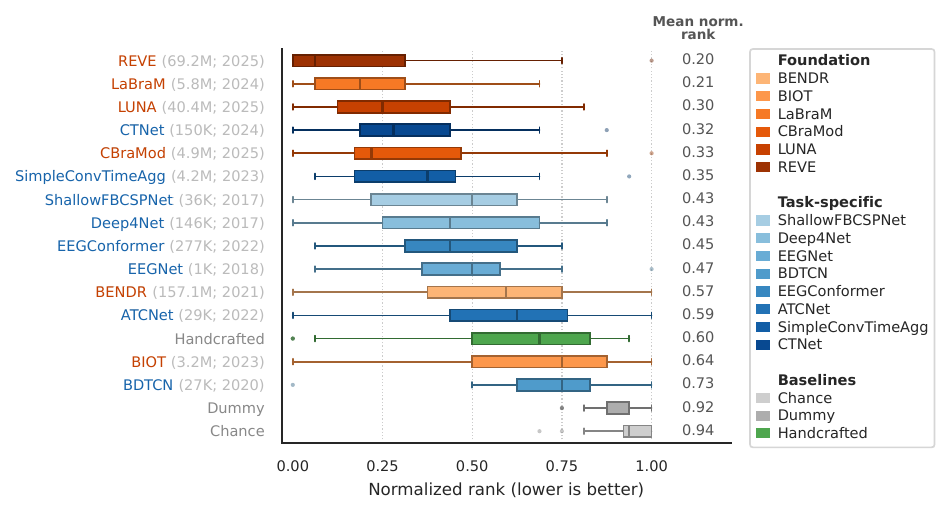}
    \caption{Model ranking on \nbecore. Each box corresponds to the rank distribution of one model across downstream tasks.
    Models are ordered by their mean normalized rank, displayed on the right hand side of the boxes.
    The number of trainable parameters (excluding the last linear projection layer) and the year of original publication are shown in parentheses next to each model name.
    Recent foundation models outrank most smaller task-specific models, suggesting large self-supervised pretraining can benefit downstream performance on a varied set of downstream tasks.
    }
    \label{fig:results_rank_plot}
\end{figure}

We summarize per-model performance in \Cref{fig:results_rank_plot}, where we order the models based on their average per-task rank.
While the highest scoring models are foundation models, \ie REVE~\citep{elouahidi2025reve},  LaBraM~\citep{jiang2024labram} and LUNA~\citep{doner2025luna}, some task-specific architectures trained \textit{from scratch} also perform well, \eg CTNet~\citep{zhao2024ctnet}, SimpleConvTimeAgg~\citep{defossez2022decoding} and Deep4Net~\citep{schirrmeister2017deep}, despite significantly lower parameter counts (\eg 150K parameters for CTNet vs. 69.2M for REVE).
This is in line with \citet{kastrati2025eeg} and \citet{wu2025adabrain} which reported similar results.
Of note, BENDR~\citep{kostas2021bendr} and BIOT~\citep{yang2023biot} did not perform as well overall despite their pretraining, which might in part be explained by the necessary addition of a channel adapter linear layer to match the expected channel montage seen during pretraining, and, for BIOT, to the use of a linear projection head rather than a large MLP head like in the original publication.

\subsection{Future-proofing evaluation through task diversity and difficulty}

\begin{figure}
    \centering
    \includegraphics[width=1.0\linewidth]{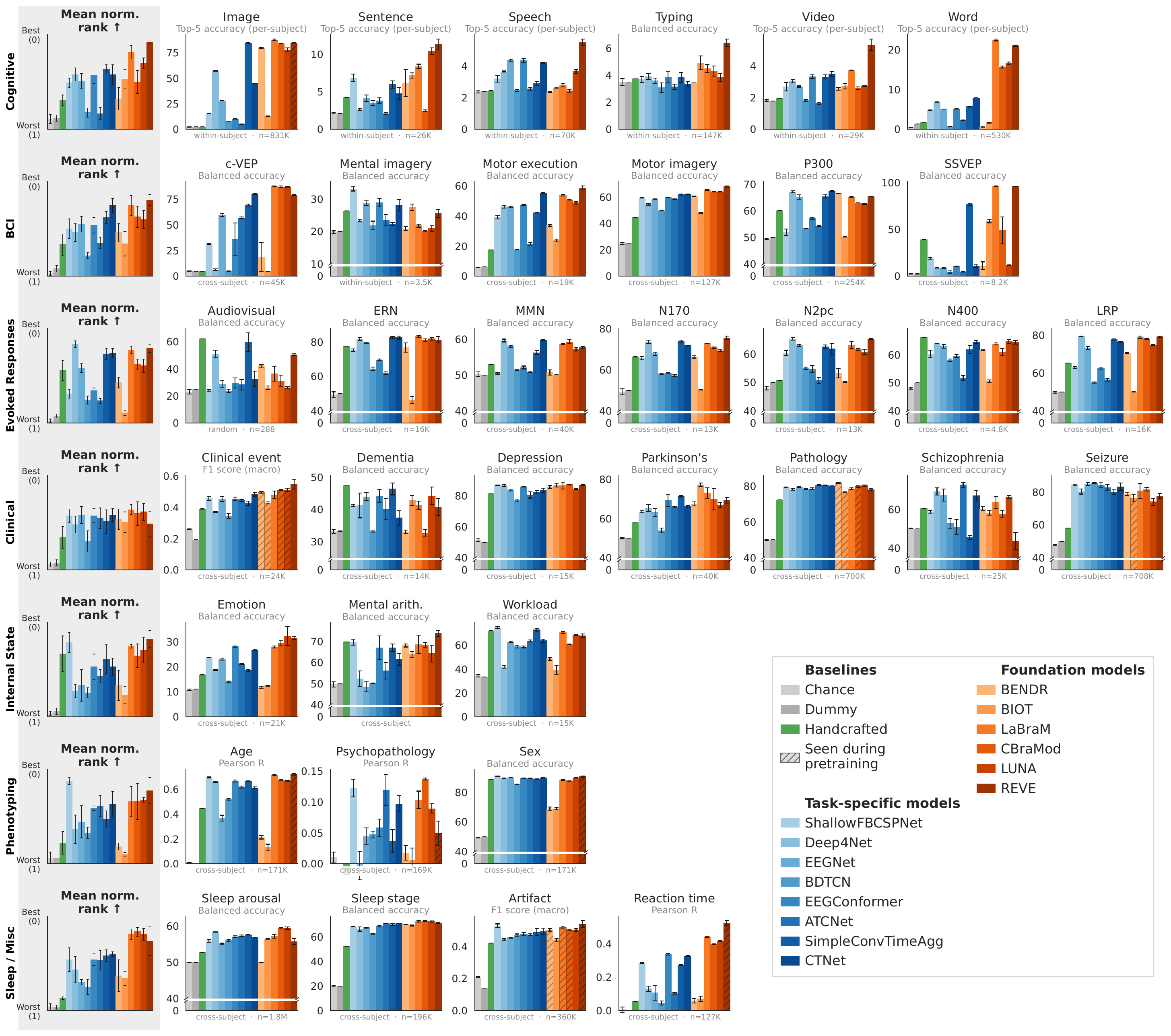}
    \caption{Performance obtained by the different task-specific (blue) and foundation (orange) models on the \ntasks downstream tasks in \nbecore. 
    Rows correspond to specific downstream task categories; the first subplot of each row summarizes the model ranking over the corresponding task category row.
    Each bar corresponds to one model, with error bars representing standard error of the mean (SEM) across seeds. Hashed bars indicate models that were pretrained on the downstream dataset, meaning the reported performance could be overinflated since test subjects or examples may have been seen by the model. 
    Dummy-level performance is evaluated for each task by predicting the majority class (classification) or the average target (regression and retrieval) computed on the training set.
    Handcrafted baselines uses sklearn pipelines which ingest covariance-like features to provide a non-deep learning baseline (see~\Cref{tab:feature_based_baselines}).
    Finally, the splitting strategy (cross-subject, within-subject or random) and the total number of examples across splits are listed under each subplot.
    }
    \label{fig:results_bar_plot}
\end{figure}

\begin{figure}
    \centering
    \includegraphics[width=1.0\linewidth]{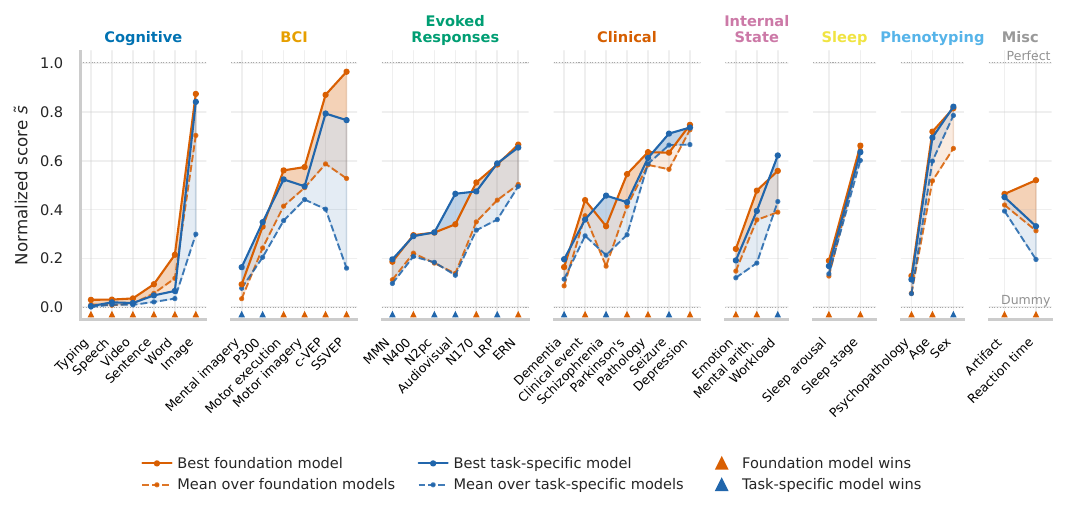}
    \caption{Normalized performance $\tilde{s}$ across downstream tasks, for the best and ``average'' foundation model (orange) and best and ``average'' task-specific models (blue). A value of 0.0 corresponds to dummy-level performance, while a value of 1.0 represents perfect (idealized) performance. 
    Horizontal bars on the x-axis indicate which model category yield the highest performance overall for a given task.
    While foundation models often outperform task-specific models, this is not the case across all tasks, \eg in evoked responses and clinical downstream tasks.
    Cognitive decoding tasks introduced in \nbe are particularly challenging, which position them well for evaluating future iterations of EEG foundation models.
    }
    \label{fig:results_agg_scores}
\end{figure}

Detailed performance results for each downstream task and model are presented in \Cref{fig:results_bar_plot} and \Cref{fig:results_agg_scores}.
The results suggest that performance over a few common EEG tasks is close to saturation, \eg \texttt{SSVEP classification}, \texttt{pathology}, \texttt{seizure} detection, \texttt{sleep stage} classification and phenotyping tasks, \ie \texttt{age} regression and \texttt{sex} classification.

In contrast, the cognitive decoding tasks are particularly challenging. These tasks, introduced in \nb (\eg \texttt{speech}, \texttt{sentence}, \texttt{video}, \texttt{word} and \texttt{image} decoding), consist of decoding dense representations of the stimulus or condition from the brain activity.
Foundation models particularly stand out on these compared to task-specific models, \eg REVE leads on the \texttt{sentence}, \texttt{speech} and \texttt{video} decoding tasks.

Models perform significantly worse on \nb's rarer paradigms, including \texttt{mental imagery}, \texttt{sleep arousal}, and \texttt{psychopathology} decoding, highlighting the difficulty of these specific decoding tasks.
Moreover, standard benchmark tasks, \ie \texttt{motor imagery}, \texttt{P300} and \texttt{N2pc} classification, become particularly challenging when framed as cross-subject tasks, often yielding performance close to dummy level.
Given the substantial margin between current results and performance ceiling, these tasks represent ideal benchmarks for future-proofing the evaluation of the next generation of EEG models.

\subsection{From \nb-\textit{Core} to \nb-\textit{Full}: Leveraging multi-study tasks for evaluating cross-dataset variability}

The results above focused on \nbecore, a version of the benchmark where each task is summarized by a single representative ``core'' dataset.
We also introduce \nbefull, an extended version of the benchmark in which downstream tasks may include additional datasets on top of their core dataset, which we leverage in two ways.

\begin{figure}
    \centering
    \includegraphics[width=1.0\linewidth]{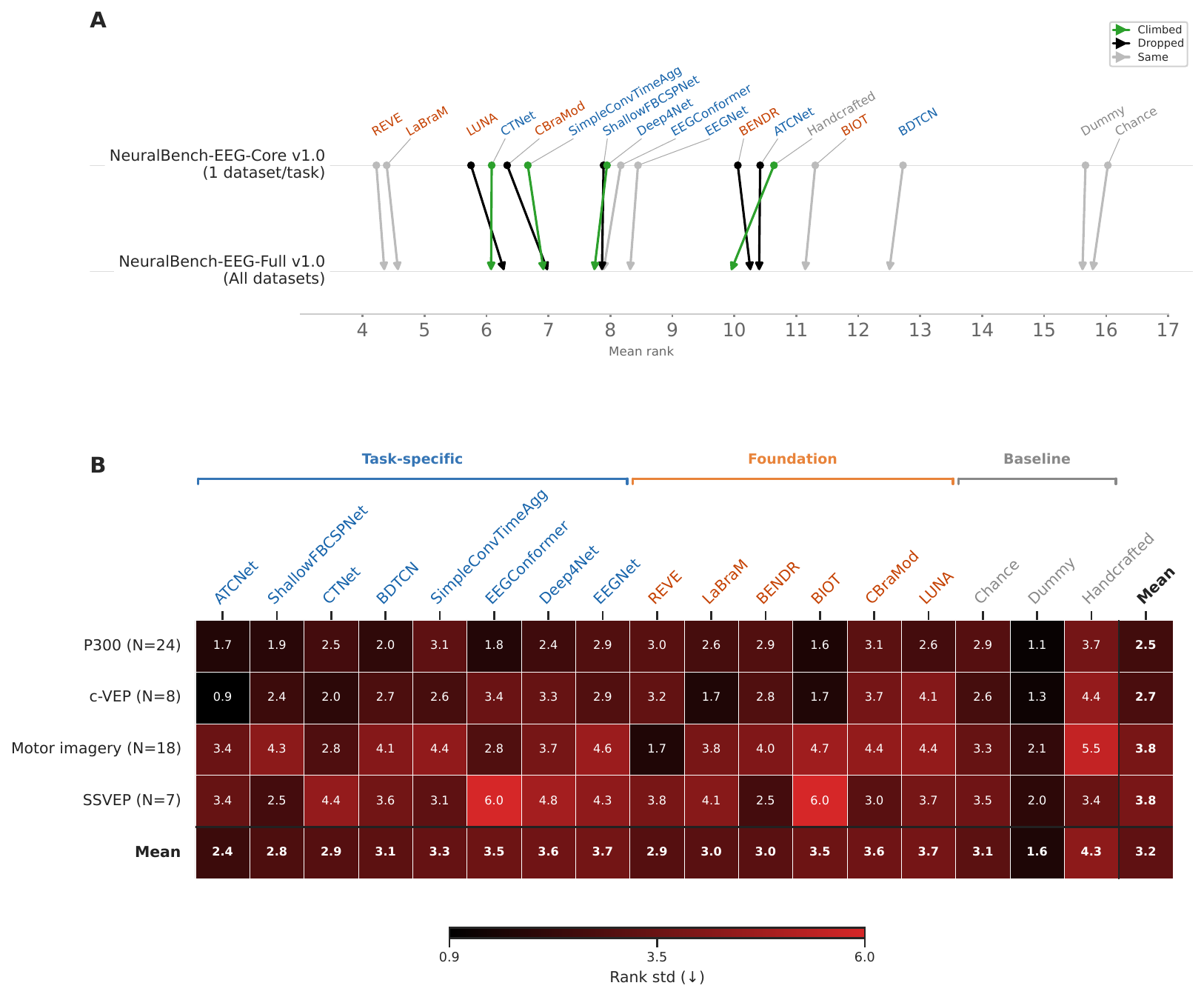}
    \caption{
    Analysing multi-dataset performance on \nbefull. 
    \textbf{A.} Difference in model ranking between the \textit{Core} and \textit{Full} variants of \nbe. Models whose global rank improve or degrade when using the full version of the benchmark are indicated by green or black lines, respectively.
    The ranking is not affected much by the addition of multiple datasets per task, which suggests \textit{Core} datasets are overall representative.
    \textbf{B.} Within-task model ranking variability. We report the standard deviation of each model's ranks across the different datasets of a task (for tasks with 5 datasets or more).
    Higher values indicate a model's performance relative to the other models vary substantially, meaning the model is not stable when evaluated across multiple datasets of the same domain.
    Tasks like motor imagery and SSVEP, and foundation models like CBraMod and LUNA, are the most unstable.
    }
    \label{fig:results_full_ranking}
\end{figure}

First, we compare models as above, but this time using dataset-averaged ranks whenever multiple datasets are available for a task, yielding a more robust ordering of models.
The ``Core'' and ``Full'' rankings are compared in \Cref{fig:results_full_ranking}A.
A Kendall's $\tau$ of 0.926 (p < 0.001) between the two rankings suggests the ``Core'' ranking is a good proxy for the ``Full'' one.
However, we notice that a few models swap positions in the ranking, \eg CTNet overtakes the LUNA foundation model as third best in the ``Full'' ranking.
This shows the gap between task-specific and foundation models is narrow enough that expanding dataset coverage is sufficient to change global ranking.

Second, we evaluate within-task variability by inspecting ranking changes in \nbefull's multi-dataset tasks (\Cref{fig:results_full_ranking}B).
REVE and LaBraM are the most stable foundation models, with the former experiencing the least rank variability of all models.
On the task dimension, we notice strong variability on \texttt{motor imagery} and \texttt{SSVEP} in particular.
For instance, on SSVEP, performance on the core dataset \citep{Wang_2017} and on \citet{Lee_2019} is mostly dominated by foundation models (LaBraM and REVE), task-specific models outperform all other models on \citet{Oikonomou2016} version A, and the handcrafted features-based baseline outperforms all or most other models on \citet{Kalunga_2016} and \citet{Oikonomou2016} versions B and C (see Supplementary~\Cref{fig:results_full_bar_plot} for details).
This strong variability suggests that multi-study comparisons are necessary to fully characterize the behavior of brain models.

Moving forward, the availability of multiple datasets per task will allow testing generalization and invariance properties of foundation models over varying recording hardware, lab environments and subject populations.
It also opens the door to multi-dataset finetuning and evaluation, \eg in which models are finetuned on a subset of task datasets, and evaluated on the remaining held out datasets.

\subsection{Extending \nb to more modalities: MEG and fMRI}

\begin{figure}
    \centering
    \includegraphics[width=0.8\linewidth]{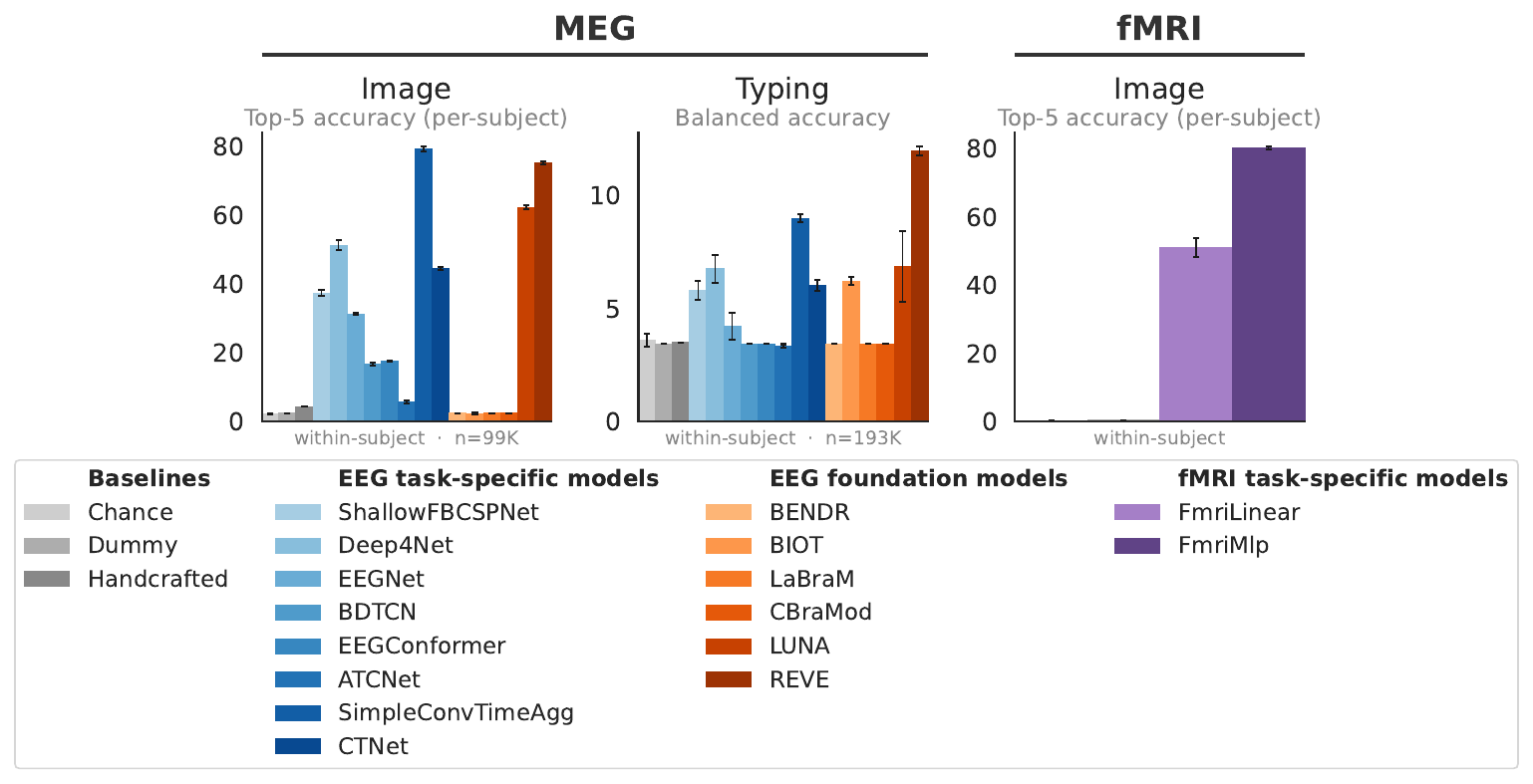}
    \caption{Performance obtained on MEG and fMRI downstream tasks.
    Each bar corresponds to one model, with error bars representing standard error of the mean (SEM) across seeds.
    See \Cref{fig:results_bar_plot} for more details.
    This first benchmark release focuses on EEG, however, as a unified framework, \nb is ready to support other brain imaging modalities as well.}
    \label{fig:results_meg_fmri}
\end{figure}

The focus of this first \nb release is EEG.
However, the benchmark infrastructure already provides support for other neuroimaging modalities, which we demonstrate by including results on two MEG and one fMRI downstream tasks (see \Cref{fig:results_meg_fmri}).

For MEG, given the similarities between the two types of data~\citep{da2013eeg} and the common reuse of EEG architectures on MEG~\citep{banville2025scaling}, we evaluate the same task-specific and foundation models as for EEG.
Expectedly, some architectures yield high performance on MEG, such as SimpleConvTimeAgg which was already shown to perform well on image decoding on the same dataset~\citep{benchetrit2024brain}.
More strikingly, some EEG foundation models, despite being pretrained on EEG data exclusively, perform well on MEG, with REVE outperforming all other models on typing decoding.
Finally, on fMRI, we present results on a similar image decoding task as in EEG and MEG, but focused on two task-specific architectures that rely on linear and convolutional layer primitives.
These results exhibit the flexibility of \nb, which is ready to be extended to other modalities as well, such as iEEG, functional near-infrared spectroscopy (fNIRS), and other neural signal types such as electromyography (EMG).

Overall, in this section, we highlighted \nb's comprehensive downstream task coverage, its versatility, allowing easy between- and within-task performance comparisons, and its extendability to more neuroimaging devices.

\section{Discussion}

\paragraph{EEG foundation vs. task-specific models.}
We introduce \nb, a unifying framework to benchmark NeuroAI models, and release \nbe, a first comprehensive benchmark evaluating existing task-specific and foundation models of EEG signals.
Our results already reveal that current EEG foundation models only marginally outperform specialized models. 
While self-supervised pretraining yields generalizable representations, downstream performance gains remain limited. 
This modest advantage may stem from the use of finetuning as a downstream strategy: in data-rich scenarios, specialized models likely possess sufficient capacity to close the performance gap with foundation models. Future evaluations should thus also incorporate evaluations in few-shot and low-data regimes (\eg 5\% of training data).

\paragraph{EEG task expansions.}
\nbe significantly expands the diversity of downstream evaluations by introducing and integrating a large set of cognitive, phenotyping, clinical and BCI ``tasks''. 
Across most of these tasks, performance remains modest, which leaves room for improvement in next generations of EEG foundation models.
In the future, we recommend maintaining a diverse set of tasks -- some easy, others difficult -- to validate the ability of models to extract valuable information from noisy brain signals, while ensuring that benchmarking is representative of downstream neuroscience and clinical applications.

\paragraph{Unique optimization strategies.}
Rather than pursuing model-specific optimization, we utilize a standardized training recipe to evaluate each architecture's ``out-of-the-box'' utility. This decision reflects a core requirement for NeuroAI foundation models: they must be easily adaptable and stable under standard configurations to be practically useful for the broader research community.

\paragraph{Unique splitting strategy.}
We used specific splitting strategies to maintain a manageable computational footprint while ensuring diverse task coverage. Although this choice limits the assessment of performance variability (as compared to a cross-validated approach), it establishes a broad baseline for evaluating foundation models. We invite the community to help incorporate stochastic partitioning in future releases to facilitate deeper statistical validation.

\paragraph{Unique finetuning strategy.}
\nb currently relies on end-to-end finetuning, which may not capture the full potential of foundation models.
Alternative strategies, such as linear probing, parameter-efficient finetuning~\citep{hu2022lora, lester2021power} and varied token aggregation methods may further improve performance and differentiate models from one another.
These dimensions should thus be studied in greater detail in future iterations of the benchmark.

\paragraph{Beyond EEG.}
While this first release focuses on EEG -- given its prominence in the brain foundation model literature~\citep{yang2025foundation} -- \nb's infrastructure based on the \ns ecosystem~\citep{king2026neuralset} is highly flexible. We demonstrate this with preliminary MEG and fMRI tasks, paving the way for a truly multimodal benchmark, and strikingly, already showing promising transfer from EEG to MEG.
However, extending to new modalities requires significant data curation and introduces device-specific challenges, such as high auto-correlation in fMRI that complicates train/test splits~\citep{poldrack2006can}.

\paragraph{Beyond decoding.}
The current benchmark is restricted to the evaluation of models which receive brain activity as input, and predict a target label or embedding, \ie \emph{decoding} models. However, \emph{encoding} models~\citep{naselaris2011encoding}, which predict brain activity from stimuli, are also actively being developed.
Existing benchmarking efforts in the neuroAI community, including machine learning competitions, are focused on encoding, \eg the Algonauts project~\citep{cichy2021algonauts,gifford2023algonauts,gifford2024algonauts} on image and video encoding in human fMRI, the Sensorium competition~\citep{willeke2022sensorium,turishcheva2024dynamic} on image and video encoding in mouse visual cortex calcium imaging, and the Brain-Score platform~\citep{schrimpf2018brain,schrimpf2020integrative} on visual and language encoding in primate electrode recordings and human fMRI.
Integrating encoding, forecasting, and denoising tasks presents unique challenges, \eg standardizing stimuli representations and evaluation metrics, but can be readily accommodated within the \nb framework.

\paragraph{Open call for community contribution.}
Building a universal benchmark for brain activity is a vast undertaking, unlikely to be achieved by a single team. We therefore open-source the \nb codebase\footnote{\nburl} and invite the community to contribute new datasets, tasks, models, and neuroimaging modalities. 

\paragraph{Towards a unified modeling of brain activity.}
Ultimately, brain foundation models have the potential to revolutionize brain-computer interfacing, neurology, and neuroscience. Moving toward the integrated processing of multimodal brain data requires rigorous, reproducible, and comprehensive evaluation. We hope \nb serves as a foundational tool to seed and support this community-wide effort.

\subsubsection*{Acknowledgments}
We thank Alexandre Gramfort, Thomas Moreau, Arnaud Delorme, Bruno Aristimunha and Pierre Guetschel for their feedback on benchmarking best practices and the NeuroAI open-source ecosystem.
We thank the maintainers and contributors of the open-source software packages on which \nb is built.
Finally, we thank the authors and participants of the datasets used in this research for their fundamental contribution to the advancement of open NeuroAI models.

\clearpage
\newpage
\bibliographystyle{assets/plainnat}
\bibliography{main}

\clearpage
\newpage
\beginappendix

% Allow reference numbers to start with the letter S in supplementary materials
\newcommand{\beginsupplement}{
    \setcounter{table}{0}
    \renewcommand{\thetable}{S\arabic{table}}%
    \setcounter{figure}{0}
    \renewcommand{\thefigure}{S\arabic{figure}}%
    \setcounter{equation}{0}
    \renewcommand{\theequation}{S\arabic{equation}}%
}
\beginsupplement

\section{Additional details on downstream datasets and tasks}
\label{app:dataset_details}

We provide the references for all datasets used in \nbe (and the MEG and fMRI extensions) in \Cref{tab:dataset_refs}.
Detailed information about the \textit{Core} datasets for each task can be found in \Cref{tab:datasets}.
We do not repeat references when multiple datasets come from the same paper.

\begin{table}[h!]
\caption{Dataset references for all \nbe tasks, grouped by category. References for the \textit{primary} dataset are listed first; additional dataset references follow.}
\label{tab:dataset_refs}
{\small
\begin{tabular}{l l p{9.5cm}}
\toprule
Category & Task & Dataset references \\
\midrule
\multirow{6}{*}{Cognitive}
  & Image & \citet{Gifford:2022}, \citet{Contier2022}, \citet{allen2022massive} \\
  & Sentence & \citet{Hollenstein2018} \\
  & Speech & \citet{brennan2019hierarchical} \\
  & Typing & \citet{levy2025brain,zhang2025thought} \\
  & Video & \citet{liu2024eeg2video} \\
  & Word & \citet{Nieuwland2018} \\
\midrule
\multirow{6}{*}{BCI}
  & c-VEP & \citet{Cabrera_Castillos_2023}, \citet{MartinezCagigal2023Checker}, \citet{MartinezCagigal2023Pary}, \citet{Thielen_2015}, \citet{Thielen_2021} \\
  & Mental imagery & \citet{Scherer_2015} \\
  & Motor execution & \citet{Srisrisawang_2024}, \citet{schirrmeister2017deep}, \citet{Ofner_2017} \\
  & Motor imagery & \citet{Stieger_2021}, \citet{schalk2009}, \citet{barachant:tel-01196752}, \citet{Cho2017}, \citet{Dornhege_2004}, \citet{Dreyer_2023}, \citet{Faller_2012}, \citet{Grosse_Wentrup_2009}, \citet{Lee_2019}, \citet{Leeb_2007}, \citet{Liu_2024}, \citet{Schwarz_2020}, \citet{Shin_2017}, \citet{Scherer_2012}, \citet{Tangermann_2012}, \citet{wei20222021}, \citet{Yi_2014}, \citet{Zhou_2016} \\
  & P300 & \citet{Schreuder_2010}, \citet{Kappenman_2021}, \citet{Acqualagna_2013}, \citet{arico2014influence}, \citet{Cattan2019Vr}, \citet{Congedo2011}, \citet{Guger_2009}, \citet{Haufe_2011}, \citet{Hoffmann_2008}, \citet{H_bner_2017}, \citet{Huebner2018}, \citet{Kojima2024A}, \citet{Kojima2024B}, \citet{Korczowski2014A}, \citet{Korczowski2014B}, \citet{Korczowski2015A}, \citet{Lee_2019}, \citet{Riccio_2013}, \citet{Romani2025}, \citet{Schaeff_2012}, \citet{H_hne_2011}, \citet{Sosulski2019}, \citet{Treder_2011}, \citet{Treder_2014}, \citet{VanVeen2019} \\
  & SSVEP & \citet{Wang_2017}, \citet{Lee_2019}, \citet{Kalunga_2016}, \citet{Nakanishi_2015}, \citet{Oikonomou2016} \\
\midrule
\multirow{7}{*}{\shortstack[l]{Evoked\\Responses}}
  & Audiovisual stimulus & \citet{Gramfort:2013} \\
  & Error-related negativity & \citet{Kappenman_2021}, \citet{Kueper:2024}, \citet{Chavarriaga_2010} \\
  & MMN & \citet{Kappenman_2021} \\
  & N170 & \citet{Kappenman_2021} \\
  & N2pc & \citet{Kappenman_2021}, \citet{Reichert_2020} \\
  & N400 & \citet{Kappenman_2021} \\
  & Lateralized readiness potential & \citet{Kappenman_2021} \\
\midrule
\multirow{7}{*}{Clinical}
  & Clinical event detection & \citet{Harati2015} \\
  & Dementia diagnosis & \citet{Miltiadous:2023} \\
  & Depression diagnosis & \citet{Mumtaz:2018} \\
  & Parkinson's diagnosis & \citet{Singh:2021} \\
  & Pathology & \citet{lopez2017automated} \\
  & Schizophrenia diagnosis & \citet{Albrecht:2019} \\
  & Seizure detection & \citet{Dan:2023} \\
\midrule
\multirow{3}{*}{\shortstack[l]{Internal\\State}}
  & Emotion recognition & \citet{Chen2023} \\
  & Mental arithmetic & \citet{Zyma2019}, \citet{Shin_2017} \\
  & Mental workload & \citet{Hinss:2022}, \citet{Hinss_2023}, \citet{Jao_2023} \\
\midrule
\multirow{2}{*}{Sleep}
  & Sleep arousal & \citet{Ghassemi:2018} \\
  & Sleep stage & \citet{Kemp2000} \\
\midrule
\multirow{3}{*}{Phenotyping}
  & Age & \citet{shirazi2024hbn} \\
  & Psychopathology & \citet{shirazi2024hbn} \\
  & Sex & \citet{shirazi2024hbn} \\
\midrule
\multirow{2}{*}{Misc}
  & Artifact detection & \citet{hamid2020temple} \\
  & Reaction time & \citet{shirazi2024hbn} \\
\bottomrule
\end{tabular}
}
\end{table}

Of note, these datasets sometimes contain stimuli presented to the participants (\eg images, videos, text, etc.).
We only make use of stimuli in the cognitive decoding tasks \citep{gifford2022large,Hollenstein2018,brennan2019hierarchical,liu2024eeg2video,Nieuwland2018}, as these tasks require the alignment of brain data to a dense representation of the stimuli.
Apart from these cases, we discard stimulus information and only keep targets (\eg \textit{classes} for classification tasks).

Detailed and up-to-date description of the downstream tasks can be found in the benchmark documentation at \nburl.

\section{Task-specific architectures and foundation models for EEG downstream tasks}

We list the task-specific deep-learning architectures, handcrafted features-based sklearn baselines, and foundation models used on the EEG downstream tasks in Tables~\ref{tab:classic_models}, \ref{tab:feature_based_baselines} and \ref{tab:foundation_models}.

\begin{table}[t]
\caption{Task-specific EEG models evaluated in \nbe.
Parameter counts are backbone-only (excluding the task-specific output head)
and are approximate as they depend on the number of input channels
(we use a representative value of $C{=}22$) here.}
\label{tab:classic_models}
\centering
\small
\begin{tabular}{l r l}
\toprule
Model & \# Params & Reference \\
\midrule
    ShallowFBCSPNet & 36K & \citet{schirrmeister2017deep} \\
    Deep4Net & 146K & \citet{schirrmeister2017deep} \\
    EEGNet & 1.5K & \citet{lawhern2018eegnet} \\
    BDTCN & 27K & \citet{gemein2020bdtcn} \\
    EEGConformer & 277K & \citet{song2022eegconformer} \\
    ATCNet & 29K & \citet{altaheri2022atcnet} \\
    SimpleConvTimeAgg & 4.2M & \citet{elouahidi2023eegsimpleconv} \\
    CTNet & 150K & \citet{zhao2024ctnet} \\
\bottomrule
\end{tabular}
\end{table}

\begin{table}[t]
\caption{Handcrafted features-based sklearn baselines evaluated in \nbe.
These pipelines are fit once on the concatenation of the training and validation sets and evaluated with the exact same preprocessing and splits as the deep learning models.
A single pipeline is used per task type; see \textit{Tasks} column.}
\label{tab:feature_based_baselines}
\centering
\small
\begin{tabular}{l p{5.7cm} p{3.8cm} p{3cm}}
\toprule
Pipeline & Description & Tasks & Reference \\
\midrule
    \texttt{XdawnTsLR} & xDAWN covariances $\rightarrow$ shrinkage $\rightarrow$ tangent space $\rightarrow$ standard scaling $\rightarrow$ logistic regression (CV-tuned $C$) & Classification: \textit{Evoked} response and \textit{P300} & \citet{rivet2009xdawn}, \citet{barachant2014plug} \\
    \texttt{CovTsLR} & sample covariances $\rightarrow$ shrinkage $\rightarrow$ tangent space $\rightarrow$ standard scaling $\rightarrow$ logistic regression (CV-tuned $C$; one-vs-rest for multilabel) & Classification: \textit{BCI} (except P300 and SSVEP), \textit{Internal State}, \textit{Clinical}, \textit{Sleep}, \textit{sex}, \textit{artifact} & \citet{barachant2011multiclass} \\
    \texttt{CoSpectraLogLR} & Welch co-spectra $\rightarrow$ upper-triangular flatten with $\log(1+x)$ on the diagonal (per-channel PSD) $\rightarrow$ standard scaling $\rightarrow$ logistic regression (CV-tuned $C$) & Classification: \textit{SSVEP} & \citet{Kalunga_2016} \\
    \texttt{CovTsRidge} & sample covariances $\rightarrow$ shrinkage $\rightarrow$ tangent space $\rightarrow$ standard scaling $\rightarrow$ ridge regression (CV-tuned $\alpha$) & Regression: \textit{age}, \textit{psychopathology}, \textit{reaction time}; Retrieval (treated as multivariate regression here): \textit{Cognitive} & \citet{barachant2011multiclass}, \citet{hoerl1970ridge} \\
\bottomrule
\end{tabular}
\end{table}

\begin{table}[t]
\caption{EEG foundation models evaluated in \nbe.
Parameter counts are backbone-only (excluding the task-specific linear probe).
Pretraining data quantities are reported as published;
TUEG refers to the Temple University Hospital EEG Corpus.}
\label{tab:foundation_models}
\centering
\small
\begin{tabular}{l p{2cm} r p{3.5cm} p{3cm} l}
\toprule
Model & Variant & \# Params & Pretraining & Data & Reference \\
\midrule
    BENDR & --- & 157.1M & Wav2vec-like contrastive & TUEG, ${\sim}$1.5K h & \citet{kostas2021bendr} \\
    BIOT & 6-datasets-18chs & 3.2M & Contrastive + supervised & 6 datasets (CHB-MIT, TUAB, TUEV, +3 private) & \citet{yang2023biot} \\
    LaBraM & Base & 5.8M & VQ-VAE masked prediction & 16 datasets, ${\sim}$2.5K h & \citet{jiang2024labram} \\
    % EEGPT & --- & 25.3M & Autoregressive reconstruction & 5 datasets, ${\sim}$200 h & \citet{wang2024eegpt} \\
    CBraMod & --- & 4.9M & Masked patch reconstruction & TUEG, ${\sim}$27K h & \citet{wang2025cbramod} \\
    LUNA & Large & 40.4M & Masked patch reconstruction & TUEG + Siena & \citet{doner2025luna} \\
    REVE & Base & 69.2M & Masked patch reconstruction & 92 datasets, ${\sim}$60K h & \citet{elouahidi2025reve} \\
\bottomrule
\end{tabular}
\end{table}

\section{Detailed splitting strategy}
\label{sec:splitting}

We split each dataset into a single training, validation and testing partition using a fixed seed.
The way we split depends on the datasets.
When datasets explicitly provide a development and evaluation split, we use it as is.
This is the case for \texttt{clinical event} classification on TUEV~\citep{Harati2015}, \texttt{pathology} detection on TUAB~\citep{lopez2015automated}, \texttt{image} decoding on THINGS-EEG2~\citep{gifford2022large} and \texttt{motor imagery} classification on the BEETL AI Challenge datasets~\citep{wei20222021}.

For cognitive decoding experiments, we focus on ``leave-concept-out'' evaluation, \ie all subjects are seen in both training and testing splits, but a common subset of the sessions or concepts are held out for testing.
% This allows testing the capacity of the model to generalize to new stimulation content, rather than to new subjects.
In many cases, this setting is required to obtain performance significantly above chance-level, as cross-subject generalization is particularly challenging on these types of tasks.
For \texttt{speech} decoding on \citet{brennan2019hierarchical}, we keep the last 4 runs of each recording as test set.
For \texttt{sentence} decoding on ZuCo v1~\citep{Hollenstein2018}, we randomly hold out 20\% of the sentences read by the subjects to use as test set.
For \texttt{word} decoding on \citet{Nieuwland2018}, we randomly hold out 10\% of the sentences read by the subjects to use as test set.
For \texttt{video} decoding on SEED-DV~\citep{liu2024eeg2video}, we keep a random 20\% of the video concepts (\eg land animal, human, natural scene, etc.) to use as test set.
For \texttt{typing} decoding on \citep{levy2025brain}, sentences are clustered using their similarity in sentence embedding space (TF-IDF), and 20\% of similar sentences are held out for testing.

Most remaining tasks/datasets use a cross-subject split.
For \texttt{age}, \texttt{psychopathology}, \texttt{sex} and \texttt{reaction time} prediction, all on the HBN-EEG dataset~\citep{shirazi2024hbn}, we leave out all recordings from release 5 (out of 11 releases) for testing, which corresponds to $\sim$329 held-out subjects.
For other tasks, we hold out a random 20\% of the subjects for testing.
Of note, this cross-subject strategy is particularly challenging for certain tasks as inter-subject variability renders generalization difficult, \eg in BCI~\citep{saha2020intra}.
Moreover, on clinical diagnosis tasks where subjects receive a single diagnosis (\texttt{dementia}, \texttt{depression}, \texttt{Parkinson's disease} and \texttt{schizophrenia}), we use stratified splits based on the diagnosis label.

A few datasets instead use a within-subject split by default to accommodate low (<10) subject counts, by keeping the last (few) sessions or runs for testing \citep{Zhou_2016,Chavarriaga_2010,Hoffmann_2008,Kueper:2024}.

Finally, we use fully random splits, \ie at the example level, for three low-sample datasets: \texttt{audiovisual stimulus} classification on the MNE Sample dataset~\citep{Gramfort:2013} and \texttt{motor imagery} classification on the datasets from \citet{Dornhege_2004,barachant:tel-01196752}.

Validation sets (used for early stopping) are sampled from the remaining data, using the same or a similar splitting strategy as used for the test set, targeting another 20\% of the overall dataset size.
We train each model on each task three times, using three different random seeds for the initialization of the (non-pretrained) model parameters.

\section{Performance metrics}
\label{sec:metrics}

We pick a single representative metric per task type to report in this document.
For binary and multiclass classification, we use the balanced accuracy:
\begin{equation}
    \text{Balanced Accuracy} = \frac{1}{C} \sum_{c=1}^{C} \frac{TP_c}{TP_c + FN_c}
\end{equation}
where $C$ is the number of classes, $TP_c$ is the number of true positives for class $c$, and $FN_c$ is the number of false negatives for class $c$.

For multilabel classification, we use the macro F1-score:
\begin{equation}
    \text{Macro F1} = \frac{1}{C} \sum_{c=1}^{C} F1_c = \frac{1}{C} \sum_{c=1}^{C} \frac{2 P_c R_c}{P_c + R_c}
\end{equation}
where $P_c = \frac{TP_c}{TP_c + FP_c}$ is the precision and $R_c = \frac{TP_c}{TP_c + FN_c}$ is the recall for class $c$.

For univariate regression, we use the Pearson correlation:
\begin{equation}
    r = \frac{\sum_{i=1}^{n} (y_i - \bar{y})(\hat{y}_i - \bar{\hat{y}})}{\sqrt{\sum_{i=1}^{n} (y_i - \bar{y})^2} \sqrt{\sum_{i=1}^{n} (\hat{y}_i - \bar{\hat{y}})^2}}
\end{equation}
where $y_i$ are the true values, $\hat{y}_i$ are the predicted values, $\bar{y}$ and $\bar{\hat{y}}$ are their respective means, and $n$ is the number of samples.

Finally, for retrieval, we use the top-5 accuracy:
\begin{equation}
    \text{Top-}k \text{ Accuracy} = \frac{1}{N} \sum_{i=1}^{N} \mathds{1}\left[ y_i \in \hat{Y}_i^{(k)} \right]
\end{equation}
with $k=5$, where $N$ is the number of queries, $y_i$ is the true label (or relevant item) for query $i$, $\hat{Y}_i^{(k)}$ is the set of the top-$k$ retrieved candidates for query $i$, and $\mathds{1}[\cdot]$ is the indicator function.

More metrics are computed in \nb, \eg root mean-squared error (RMSE) for regression and median ranks for retrieval, but are not reported here for brevity.

\section{Additional results on \nbecore}

\begin{figure}
    \centering
    \includegraphics[width=1.0\linewidth]{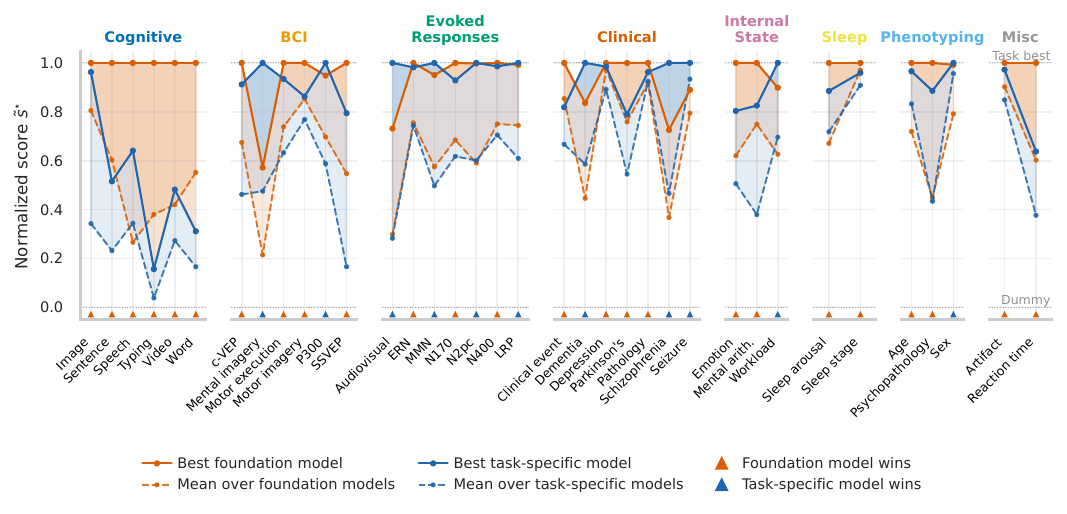}
    \caption{Max-normalized performance $\tilde{s}^*$ across downstream tasks, for the best and ``average'' models. As in \Cref{fig:results_agg_scores}, but scores are normalized to be between ``dummy'' performance and per-task best model performance, highlighting variability as compared to the highest performing model.
    }
    \label{fig:results_agg_scores_best}
\end{figure}

\Cref{fig:results_agg_scores_best} shows max-normalized scores over core datasets.

\section{Additional results on \nbefull}

\begin{figure}
    \centering
    \includegraphics[width=1.0\linewidth]{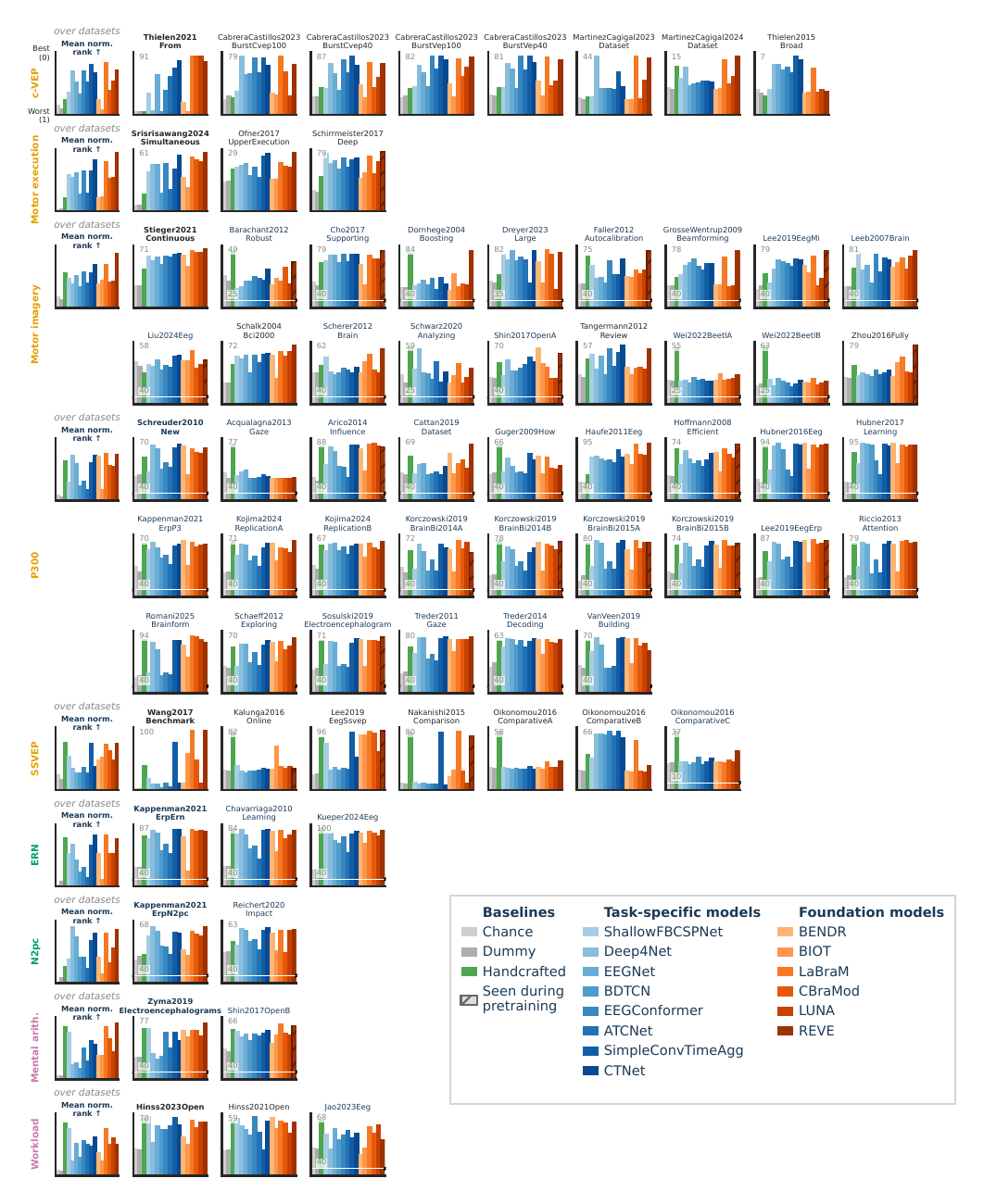}
    \caption{Performance obtained by the different task-specific (blue) and foundation (orange) models on the multi-dataset \nbefull downstream tasks (see~\Cref{fig:results_bar_plot}'s caption for details).
    }
    \label{fig:results_full_bar_plot}
\end{figure}

\Cref{fig:results_full_bar_plot} shows performance on each multi-dataset downstream task included in \nbefull.

\end{document}

%% file: math_commands.tex
\usepackage{amsmath,amsfonts,bm}

\def\eqref#1{equation~\ref{#1}}
\def\1{\bm{1}}

\DeclareMathAlphabet{\mathsfit}{\encodingdefault}{\sfdefault}{m}{sl}
\SetMathAlphabet{\mathsfit}{bold}{\encodingdefault}{\sfdefault}{bx}{n}

\newcommand{\ie}{\textit{i}.\textit{e}., }
\newcommand{\eg}{\textit{e}.\textit{g}.\ }